\documentclass[letterpaper, 10 pt, conference]{ieeeconf}  

\IEEEoverridecommandlockouts                              

\usepackage{cite}
\usepackage{svg}
\usepackage{amsmath,amssymb,amsfonts}
\usepackage{algorithmic}
\usepackage{graphicx}
\usepackage{graphics}
\usepackage{textcomp}
\usepackage{xcolor}
\usepackage{dsfont}
\usepackage{subfigure}
\usepackage{url}
\usepackage{tabularx,booktabs}
\usepackage{multirow}
\usepackage{bm}
\usepackage{balance}

\makeatletter
\let\NAT@parse\undefined
\makeatother
\usepackage{hyperref}

\title{\LARGE \bf
ADAPT: Action-aware Driving Caption Transformer
}

\author{Bu Jin$^{1,2}$, Xinyu Liu$^{2,5}$, Yupeng Zheng$^{1,2}$, Pengfei Li$^{3}$, Hao Zhao$^{4}$, Tong Zhang$^{6}$, \\ Yuhang Zheng$^{2,7}$, Guyue Zhou$^{2}$ and Jingjing Liu$^{2}$
\thanks{$^{1}$Institute of Automation, Chinese Academy of Sciences, China,
        \{jinbu2022, zhengyupeng2022\}@ia.ac.cn.}%
\thanks{$^{2}$Institute for AI Industry Research (AIR), Tsinghua University, China,
        \{zhouguyue, JJLiu\}@air.tsinghua.edu.cn.}%
\thanks{$^{3}$Department of Computer Science and Technology, Tsinghua University, China,
        li-pf22@mails.tsinghua.edu.cn.}%
\thanks{$^{4}$Intel Labs China, Peking University, China,  
        zhao-hao@pku.edu.cn, hao.zhao@intel.com.}
\thanks{$^{5}$Xidian University, China,
        liuxinyu@stu.xidian.edu.cn.}
\thanks{$^{6}$Southern University of Science and Technology, China,  
        11911611@mail.sustech.edu.cn.}
\thanks{$^{7}$School of Mechanical Engineering and Automation, Beihang University, China,  
        zyh\_021@buaa.edu.cn.}
}

\begin{document}

\maketitle

\begin{abstract}
End-to-end autonomous driving has great potential in the transportation industry. 
However, the lack of transparency and interpretability of the automatic decision-making process hinders its industrial adoption in practice. 
There have been some early attempts to use attention maps or cost volume for better model explainability which is difficult for ordinary passengers to understand. 
To bridge the gap, we propose an end-to-end transformer-based architecture, ADAPT (Action-aware Driving cAPtion Transformer), which provides \emph{user-friendly} natural language narrations and reasoning for each decision making step of autonomous vehicular control and action. 
ADAPT jointly trains both the driving caption task and the vehicular control prediction task, through a shared video representation. 
Experiments on BDD-X (Berkeley DeepDrive eXplanation) dataset demonstrate state-of-the-art performance of the ADAPT framework on both automatic metrics and human evaluation. 
To illustrate the feasibility of the proposed framework in real-world applications, we build a novel deployable system that takes raw car videos as input and outputs the action narrations and reasoning in real time. The code, models and data are available at https://github.com/jxbbb/ADAPT.

\end{abstract}

\section{Introduction}


The goal of an autonomous system is to gain precise perception of the environment, make safe real-time decisions, take reliable actions without human involvement and provide a safe and comfortable ride experience for passengers. 
There are generally two types of paradigms for autopilot controller design: mediation-aware method \cite{behere2015functional, yurtsever2020survey} and end-to-end learning approach \cite{sauer2018conditional, xiao2020action, chen2021learning, toromanoff2020end, wang2020learning, bojarski2016end, codevilla2018end, muller2018driving, pomerleau1988alvinn, wei2021perceive, zeng2019end, casas2021mp3, chen2020learning, filos2020can, behl2020label, buhler2020driving, codevilla2019exploring, huang2020multi, ohn2020learning, prakash2020exploring, xiao2020multimodal}. Mediation-aware approaches rely on recognizing human-specified features such as vehicles, lane markings, etc., which require rigorous parameter tuning to achieve satisfactory performance.
In contrast, end-to-end methods directly take raw data from sensors as input to generate planning routes or control signals.

One of the key challenges in deploying such autonomous control systems to real vehicles is that intelligent decision-making policies in autonomous cars are often too complicated and difficult for common passengers to understand, for whom the safety of such vehicles and their controlability is the top priority.

Some previous work has explored the interpretation of autonomous navigation \cite{kim2017interpretable, hu2021fiery, philion2020lift, wang2021learning, cui2021lookout, zeng2019end, zeng2020dsdnet, sadat2020perceive, casas2021mp3}. 
Cost map, for example, is employed in \cite{zeng2019end} to interpret the actions of a self-driving system by visualizing the difficulty of traversing through different areas of the map. 
Visual attention is utilized in \cite{kim2017interpretable} to filter out non-salient image regions, and \cite{saha2022translating} constructs BEV (Bird’s eye view) to visualize the motion information of the vehicle. 
However, these interfaces can easily lead to misinterpretation if the user is unfamiliar with the system.

\begin{figure}
  \centering
  \includegraphics[width=0.45\textwidth]{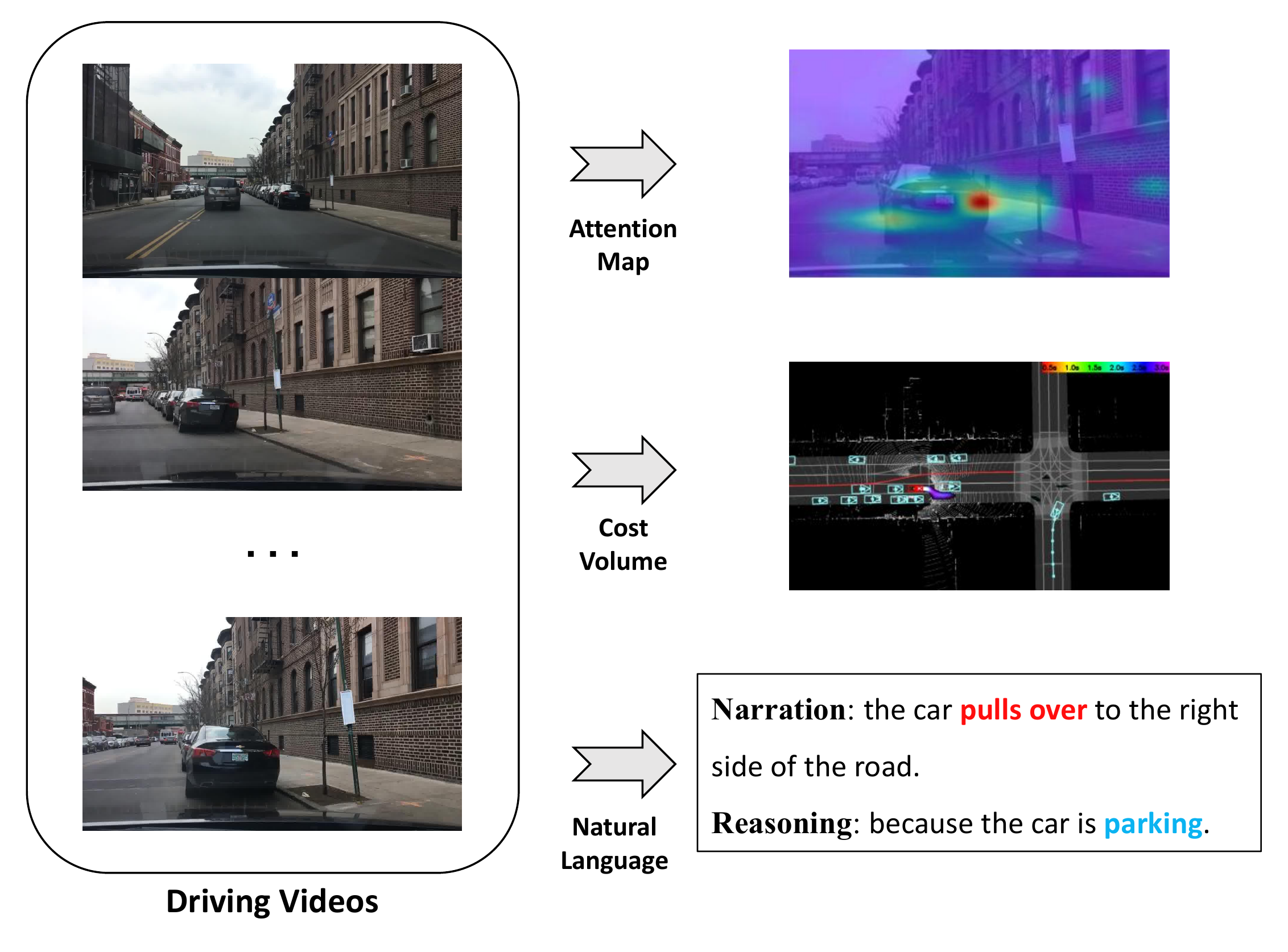}
  \caption{Different interpretation methods of self-driving vehicles, including attention map\cite{kim2017interpretable}, cost volume\cite{zeng2019end} and natural language. Despite the effectiveness of attention map or cost volume, language-based interpretation is more user-friendly to common passengers.}
  \label{fig:teaser}
  \vspace{-0.2cm}
\end{figure}

An ideal solution is to include natural language narrations to guide the use throughout the decision making and action taking process of the autonomous control module, which is comprehensible and user-friendly. Furthermore, an additional reasoning explanation for each control/action decision can help users understand the current state of the vehicle and the surrounding environment, as supporting evidence for the actions taken by the autonomous vehicle. For example, "[\textit{Action narration}:] the car pulls over to the right side of the road, [\textit{Reasoning}:] because the car is parking", as shown in Fig.~\ref{fig:teaser}.
Explaining vehicle behaviors via natural language narrations and reasoning thus makes the whole autonomous system more transparent and easier to understand.

To this end, we propose ADAPT, the first action-aware transformer-based driving action captioning architecture that provides for passengers user-friendly natural language narrations and reasoning of autonomous driving vehicles. 
To eliminate the discrepancy between the captioning task and the vehicular control signal prediction task, we jointly train these two tasks with a shared video representation. This multi-task framework can be built upon various end-to-end autonomous systems by incorporating a text generation head.

We demonstrate the effectiveness of the ADAPT approach on a large-scale dataset that consists of control signals and videos along with action narration and reasoning. 
Based on ADAPT, we build a novel deployable system that takes raw vehicular navigation videos as input and generates the action narrations and reasoning explanations in real time. 

Our contributions can be summarized as:

\begin{itemize}
\item[$\bullet$] We propose ADAPT, a new end-to-end transformer-based action narration and reasoning framework for self-driving vehicles. 
\item[$\bullet$] We propose a multi-task joint training framework that aligns both the driving action captioning task and the control signal prediction task. 
\item[$\bullet$] We develop a deployable pipeline for the application of ADAPT in both the simulator environment and the real world.
\end{itemize}

\begin{figure*}
  \centering
  \includegraphics[width=1.0\textwidth]{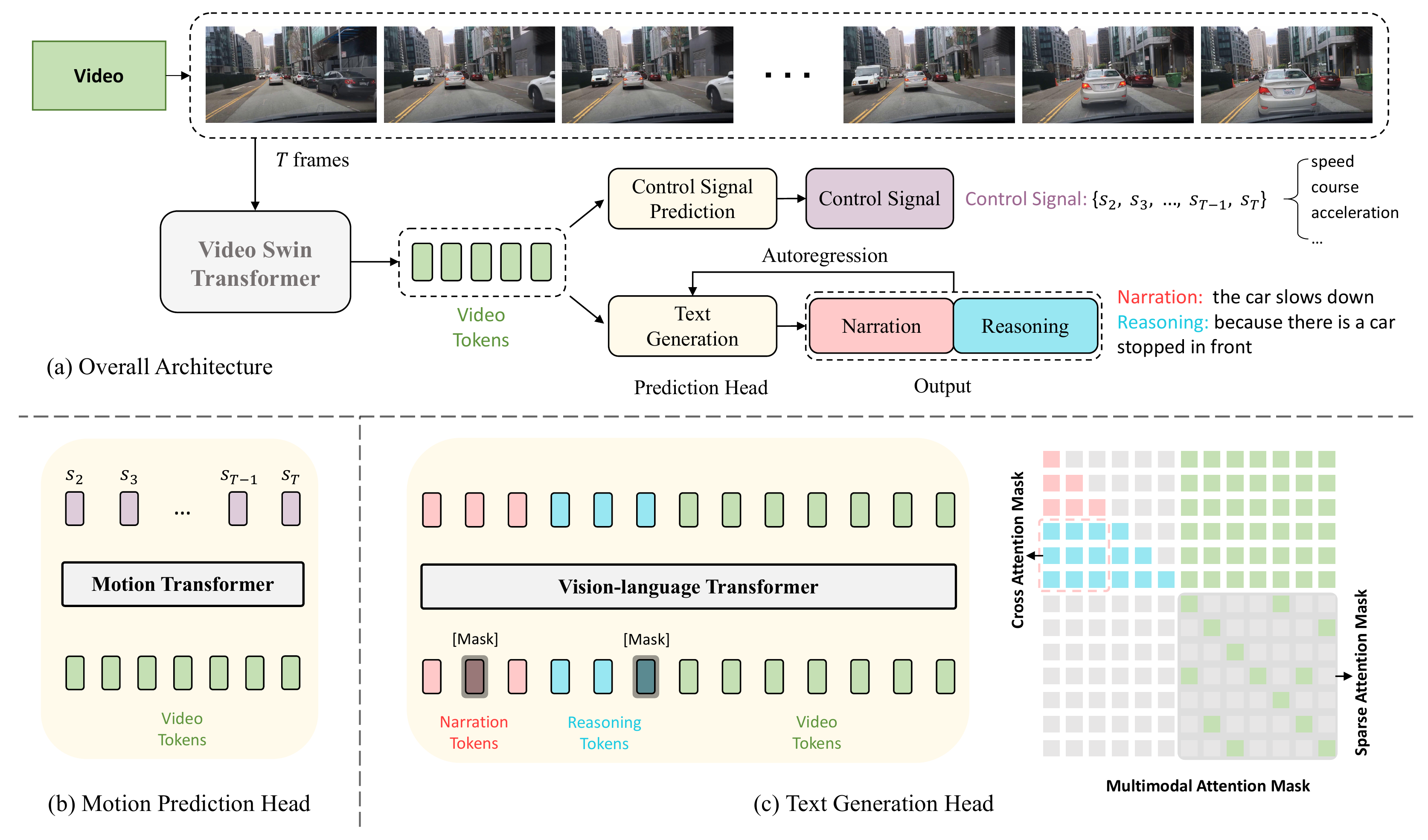}
  \caption{Overview of ADAPT framework. (a) Input is a vehicle-front-view video, and outputs are predicted vehicle's control signals and the narration and reasoning of the current action. We first densely and uniformly sample $T$ frames from the video, which are sent to the learnable video swin transformer and tokenized into video tokens. Different prediction heads generate final motion results and text results. (b)(c) show the prediction heads, respectively.}
  \label{fig:main}
  \vspace{-0.2cm}
\end{figure*}

\section{Related Work}

\subsection{Video Captioning}

The main goal of the video captioning task is to describe the objects and their relationship of a given video in natural language.
Early researches \cite{guadarrama2013youtube2text, hanckmann2012automated, kojima2002natural, rohrbach2013translating} generate sentences with specific syntactic structures by filling recognized elements in fixed templates, which are inflexible and lack of richness. 
\cite{chen2017video, gao2022hierarchical, ji2021multi, pan2016jointly, pan2017video, pasunuru2017multi, venugopalan2015sequence, venugopalan2014translating, yao2015describing, yu2016video} exploit sequence learning approaches to generate natural sentences with flexible syntactic structures.
Specifically, these methods employ a video encoder to extract frame features and a language decoder to learn visual-textual alignment for caption generation.
To enrich captions with fine-grained objects and actions, \cite{zhao2018video, pan2020spatio, zhang2020object} exploit object-level representations that capture detailed object-aware interaction features in videos.
\cite{liu2018sibnet} further develops a novel dual-branch convolutional encoder to jointly learn the content and semantic information of videos. 
Moreover, \cite{jin2020sbat} adapts the uni-modal transformer to video captioning and employs a sparse boundary-aware pooling to reduce the redundancy in video frames. The development of scene understanding \cite{wang2022image, he2022masked, liu2021swin, zhang2022mutr3d, han2022scene, huang2023diffusion, li2022distance, jin2022language, li2022toist, chen2022pq, zhao2020learning, zhao2017physics} also contribute a lot to the captioning task.
Most recently, \cite{lin2022swinbert} proposes an end-to-end transformer-based model SWINBERT, which utilizes a sparse attention mask to lessen the redundant and irrelevant information in consecutive video frames.

While existing architectures achieve promising results for general video captioning, it cannot be directly applied to action representation because simply transferring video caption to self-driving action representation would miss some key information like the speed of the vehicle, which is essential in the autonomous system. How to effectively use these multi-modal information to generate sentences remains a mystery, which is the focus of our work.

\subsection{End-to-End Autonomous Driving}
Learning-based autonomous driving is an active research area \cite{janai2020computer, tampuu2020survey}. 
Some learning-based driving methods such as affordances \cite{sauer2018conditional, xiao2020action} and reinforcement learning \cite{chen2021learning, toromanoff2020end, wang2020learning} are employed, gaining promising performance. 
Imitation methods \cite{bojarski2016end, codevilla2018end, muller2018driving, pomerleau1988alvinn, wei2021perceive, zeng2019end} are also utilized to regress the control commands from human demonstrations. 
For example, \cite{casas2021mp3, chen2020learning, filos2020can} model the future behavior of driving agents like vehicles, cyclists or pedestrians to predict the vehicular waypoints, while \cite{behl2020label, buhler2020driving, codevilla2019exploring, huang2020multi, ohn2020learning, prakash2020exploring, xiao2020multimodal} predict vehicular control signals directly according to the sensor input, which is similar to our control signal prediction sub-task.

\subsection{Interpretability of Autonomous Driving}
Interpretability, or the ability to provide a comprehensive explanation plays a significant role in the social acceptance of artificial intelligence \cite{yuan2022situ, edmonds2019tale} and autonomous driving is no exception.
Most interpretable approaches of autonomous vehicles are vision-based \cite{saha2022translating, hu2021fiery, philion2020lift, wang2021learning, kim2017interpretable} or LiDAR-based \cite{cui2021lookout, zeng2019end, zeng2020dsdnet, sadat2020perceive, casas2021mp3}.
\cite{kim2017interpretable} first utilizes the visualization of an attention map that filters out non-salient image regions to make  autonomous vehicles reasonable and interpretable. Nevertheless, the attention map may easily include some less important areas which cause misunderstanding for passengers. 
\cite{saha2022translating, hu2021fiery, philion2020lift, wang2021learning} constructs BEV (Bird’s eye view) from a vehicle camera to visualize the motion information and environmental status of the vehicle.
\cite{zeng2019end} takes as input LiDAR and HD maps to forecast the bounding boxes of driving agents and exploits cost volume to explain the reason for the planner's decision. 
Furthermore, \cite{casas2021mp3} constructs an online map from segmentation as well as the states of driving agents to avoid heavy dependence on HD maps.

Although the vision-based or LiDAR-based approaches provide promising results, the lack of linguistic interpretation makes them too complicated for passengers like the elderly to understand.
\cite{kim2018textual} first explores the possibility of textual explanations for self-driving vehicles, which offline extracts video features from control signal prediction task and conducts video captioning afterwards. Unfortunately, the discrepancy between these two tasks makes the offline-extracted features sub-optimal for downstream captioning task, which is the focus of our work.

\subsection{Multi-task Learning in Autonomous Driving}
Our end-to-end framework adopts multi-task learning, where we train the model on a joint objective of text generation and control signal prediction. Multi-task learning helps extract more useful information by exploiting inductive biases between different tasks \cite{chen2022cerberus} and has shown promising prospects in autonomous driving. \cite{feichtenhofer2017detect, frossard2019deepsignals} shows that detection and tracking can be trained together. \cite{luo2018fast} further applies a joint detector and trajectory predictor into a single model and gains promising results. This idea is extended by \cite{casas2018intentnet} to simultaneously predict the intention of actors. More recently, \cite{zeng2019end} further includes a cost map based control signal planner in the joint model. These works show that joint training of different tasks improves the performance of individual tasks due to better data utilization and shared features, which inspires our joint training strategy of control signal prediction task and text generation task.

\section{Method}

\subsection{Overview}\label{subsec:Overview}

The ADAPT architecture is illustrated in Fig.~\ref{fig:main}, which addresses two tasks: Driving Caption Generation (DCG) and  Control Signal Prediction (CSP). 
DCG takes a sequence of raw video frames as inputs, and outputs two natural language sentences: one describes the vehicle's action (e.g., "the car is accelerating"), and the other explains the reasoning for taking this action (e.g., "because the traffic lights turn green"). 
CSP takes the same video frames as inputs, and outputs a sequence of control signals, such as speed, course or acceleration. 

Generally, DCG and CSP tasks share the same Video Encoder, while employing different prediction heads to produce the final prediction results. For DCG task, we employ a vision-language transformer encoder to generate two natural language sentences via sequence-to-sequence generation. For CSP task, we use a motion transformer encoder to predict the control signal sequence. 

\subsection{Video Encoder}\label{subsec: Video Encoder}
Following Swinbert \cite{lin2022swinbert}, we employ Video Swin Transformer (video swin) \cite{liu2022video} as the visual encoder to encode video frames into video feature tokens. 
Given a car video captured from the first perspective, we first do uniform sampling to get $T$ frames of size $ H \times W \times 3$. These frames are passed as inputs to video swin, resulting in feature $F_V$ of size $\frac{T}{2} \times \frac{H}{32} \times \frac{W}{32} \times 8C$, where $C$ is the channel dimension defined in video swin. 
Then the video features are fed into different prediction heads for individual tasks.

\subsection{Prediction Heads}

\textbf{Text Generation Head}
The purpose of the text generation head is to generate two sentences that describe both the action of the vehicle and the reason behind it. 
As mentioned in Sec.~\ref{subsec: Video Encoder}, the video frames are encoded to video features $F_V$ of size $\frac{T}{2} \times \frac{H}{32} \times \frac{W}{32} \times 8C$. Then we tokenize the video features along the channel dimension, resulting in $\frac{T}{2} \times \frac{H}{32} \times \frac{W}{32}$ tokens with dimension of $8C$.
As for the text inputs (action narrations and reasoning), we first tokenize each sentence and pad it to a fixed length. Then we concatenate these two sentences and embed them with an embedding layer. To identify the difference between action narration and reasoning, we exploit a segment embedding method (widely used in Bert\cite{devlin2018bert}) to distinguish them. And we use a learnable MLP that transforms the dimension of video tokens to ensure the dimension consistency between video tokens and text tokens.
Finally, the text tokens and video tokens are fed into the vision-language transformer encoder, which will generate a new sequence includes both action narrations and reasoning.

\textbf{Control Signal Prediction Head}
The goal of CSP head is to predict the control signals (e.g. acceleration) of the vehicle based on video frames. Given video features of $T$ frames, along with the corresponding control signal recordings $\bm{S}=\{\bm{s}_1, \bm{s}_2, .., \bm{s}_T\}$, the output of CSP head is a sequence of control signals $\hat{\bm{S}}=\{ \hat{\bm{s}_2}, ..., \hat{\bm{s}_T} \}$. Each control signal $\bm{s}_i$ or $\hat{\bm{s}_i}$ is a n-tuple, where $n$ refers to how many types of sensor we exploit. 
We first tokenize the video features, then utilize another transformer (motion transformer) to generate the prediction of these control signals. The loss function $\mathcal{L}_{CSP}$ is defined as the mean squared error of $\bm{S}$ and $\hat{\bm{S}}$:
\begin{equation}
    \mathcal{L}_{CSP}=\frac{1}{T-1} \sum_{i=2}^T \left(s_i-\hat{s}_i\right)^2
\end{equation}
Note that we do not predict control signal corresponding to the first frame, since the dynamic information of the first frame is limited, while other signals can be easily inferred from previous frames.

\subsection{Joint Training}
In our framework, we assume that CSP and DCG tasks are aligned on the semantic level of the video representation. Intuitively, action narration and the control signal data are different expression forms of the action of self-driving vehicles, while reasoning explanation concentrates on the elements of the environment that influence the action of the vehicles. We believe that jointly training these tasks in a single network can improve performance by leveraging the inductive biases between different tasks. 

During training, CSP and DCG are performed jointly. We simply add the $\mathcal{L}_{CSP}$ and $\mathcal{L}_{DCG}$ to get the final loss function:
\begin{equation}
    \mathcal{L}=\mathcal{L}_{CSP} + \mathcal{L}_{DCG}
\end{equation}

Despite the joint training of both tasks, inference on each task can be carried out independently. For the DCG task, ADAPT takes a video sequence as input, and outputs the driving caption with two segments. Text generation is performed in an auto-regressive manner. Specifically, our model starts with a "[CLS]" token and generates one word token at a time, consuming previously generated tokens as the inputs of the vision-language transformer encoder. Generation continues until the model outputs the ending token "[SEP]" or reaches the maximum length threshold of a single sentence. After padding the first sentence to the maximum length, we concatenate another "[CLS]" to the inputs and repeat the aforementioned process.

\section{Experiment}

In this section, we evaluate ADAPT over metrics of the standard captioning task, including BLEU4 \cite{papineni2002bleu}, METEOR \cite{banerjee2005meteor}, ROUGE-L \cite{lin2004automatic} and CIDEr \cite{vedantam2015cider} (abbreviated as B4, M, R and C in later tables). As quantitative evaluation of captioning is still an open question, we also provide detailed human evaluation results for the subjective correctness of the generated text. Ablation studies further demonstrate the effectiveness of the proposed joint-training 
framework. 

\subsection{Dataset}
BDD-X \cite{kim2018textual} is a driving-domain caption dataset, consisting of nearly 7000 videos paired with control signals. The videos and control signals are collected from BDD100K dataset \cite{yu2020bdd100k}.
Each video has a duration of 40 seconds on average, with 1280×720 resolution and 30 FPS. 
Each video contains 1 to 5 vehicle behaviors, such as accelerating, turning right or merging lanes. All these behaviors are accompanied by text annotation, including action narration (e.g., "the car stops") and reasoning (e.g., "because the traffic light is red"). 
There are around 29000 behavior-annotation pairs in total. 
To the best of our knowledge, BDD-X is the only driving-domain caption dataset accompanied by car videos and control signals.

\begin{table}[]
\begin{center} 
\caption{Comparison with SOTA on Video Captioning Metrics}
\label{tab:comparison}
\begin{tabular}{@{}ccccccc@{}}
\toprule
\multirow{2}{*}{Method}      & \multicolumn{3}{c}{Narration} & \multicolumn{3}{c}{Reasoning} \\ \cmidrule(l){2-4} \cmidrule(l){5-7}
                             & B4     & C    & M   & B4     & C    & M   \\ \midrule
\multicolumn{1}{c}{S2VT\cite{venugopalan2015sequence}}   & 30.2  & 179.8    & 27.5   & 6.3    & 53.4    & 11.2   \\
\multicolumn{1}{c}{S2VT++\cite{venugopalan2015sequence}} & 27.1   & 157.0    & 26.4   & 5.8    & 52.7    & 10.9   \\
\multicolumn{1}{c}{SAA\cite{kim2018textual}} & 31.8    & 214.8    & 29.1   & 7.1    & 66.1    & 12.2   \\
\multicolumn{1}{c}{WAA\cite{kim2018textual}} & 32.3    & 215.8    & 29.2   & 7.3    & 69.5    & 12.2   \\
\multicolumn{1}{c}{Ours}     &   
  \textbf{34.6} &
  \textbf{247.5} &
  \textbf{30.6} &
  \textbf{11.4} &
  \textbf{102.6} &
  \textbf{15.2} \\ \bottomrule
\end{tabular}
\end{center}
\vspace{-0.2cm}
\end{table}

\begin{table}[]
\begin{center} 
\caption{Comparison with SOTA on Human Evaluation}
\label{tab:human eval}
\begin{tabular}{@{}cccc@{}}
\toprule
Method   & Narration  & Reasoning   & Full sentence \\ \midrule
SAA\cite{kim2018textual}   & 90.8\%        & 62.4\%        & -              \\
WAA\cite{kim2018textual}   & \textbf{93.5}\%        & 66.0\%        & -              \\
Ours                    & 90.0\%        & \textbf{90.3}\%        & 82.7\%              \\ \bottomrule
\end{tabular}
\end{center}
\vspace{-0.5cm}
\end{table}

\subsection{Implementation Details}
The video swin transformer is pre-trained on Kinetics-600 \cite{carreira2018short}, while the vision-language transformer and motion transformer are randomly initialized. Note that in our implementation we do not freeze the parameters of video swin, so ADAPT is trained in a complete end-to-end manner. 
The input video frames are resized and cropped to the spatial size of 224. And for narration and reasoning, we use WordPiece embeddings
\cite{devlin2018bert} instead of the whole words (e.g., "stops" is cut to "stop" and "\#s") and the maximal length of each sentence is 15. 
During training period, we randomly mask $50\%$ of the tokens for masked language modeling. And the masked token has $80\%$ chance to be a "[MASK]" token, $10\%$ chance to be a random word, and $10\%$ chance to remain the same. 
We employ AdamW optimizer and use a learning rate warm-up during the early $10\%$ training steps followed by linear decay.
The whole training process for 40 epochs takes about 13 hours on 4 NVIDIA V100 GPUs with a batch size of 4 per GPU.

\subsection{Main Results}
We compare ADAPT with state-of-the-art methods on BDD-X dataset. Table~\ref{tab:comparison} shows the comparison results on standard captioning metrics. We observe that ADAPT achieves significant performance gain over existing methods. Specifically, ADAPT outperforms prior state-of-the-art work \cite{kim2018textual} by 31.7 for action narration and 33.1 for reasoning on CIDEr metric. 

In addition to automatic evaluation measures, we also conduct human evaluation to measure the subjective correctness of output narration and reasoning. The whole evaluation process is divided into three sections: (1) narration, (2) reasoning, and (3) full sentence. During the first section, a human evaluator judges whether the predicted narrations conform to the vehicle's action. In the second section, we display both ground-truth narration and predicted reasoning, and require human evaluators to judge whether the reasoning is correct. Then in the last section, both predicted narrations and predicted reasoning are displayed. Table~\ref{tab:human eval} shows that ADAPT outperforms previous work in reasoning accuracy while maintaining high accuracy on narration evaluation, demonstrating the effectiveness of ADAPT.

\begin{table}[]
\begin{center}
\caption{Single Captioning vs. Action-aware Captioning}
\label{tab:multitask}
\resizebox{1\columnwidth}{!}{
\begin{tabular}{@{}ccccccccc@{}}
\toprule
\multirow{2}{*}{Method} & \multicolumn{4}{c}{Narration}            & \multicolumn{4}{c}{Reasoning}            \\ \cmidrule(l){2-5} \cmidrule(l){6-9} 
                        & B4  & C  & M & R  & B4  & C  & M & R  \\ \midrule
Single      & 33.2 & 238.9 & 29.7 & 62.0 & 8.6 & 89.7 & 14.1 & 31.4 \\
Single+     & 33.9 & \textbf{248.3} & 30.5 & \textbf{63.1} & 9.3 & 97.2 & 14.6 & 31.5 \\
Ours &
  \textbf{34.6} &
  247.5 &
  \textbf{30.6} &
  62.8 &
  \textbf{11.4} &
  \textbf{102.6} &
  \textbf{15.2} &
  \textbf{32.0} \\ \bottomrule
\end{tabular}}
\end{center}
\vspace{-0.2cm}
\end{table}

\begin{table}[]
\begin{center} 
\caption{Analysis on Control Signal Types}
\label{tab:signal type}
\begin{tabular}{@{}cccccccc@{}}
\toprule
\multicolumn{2}{c}{Signals} &
  \multicolumn{3}{c}{Narration} &
  \multicolumn{3}{c}{Reasoning} \\ \cmidrule(l){1-2} \cmidrule(l){3-5} \cmidrule(l){6-8}
    Speed &
    Course &
    C &
    M &
    R &
    C &
    M &
    R \\ \midrule
\checkmark  &               & 232.0  & 29.9 & 61.5  & 88.0  & 15.1 & 31.0  \\
            & \checkmark    & 218.2  & 29.3 & 61.2  & 88.6  & 14.1 & 30.6  \\ 
\checkmark  & \checkmark    & \textbf{247.5} & \textbf{30.6} & \textbf{62.8}  & \textbf{102.6}  & \textbf{15.2} & \textbf{32.0} \\ \bottomrule

\end{tabular}
\end{center} 
\vspace{-0.4cm}
\end{table}

\begin{table}[]
\vspace{-0.2cm}
\begin{center}
\caption{Impact of Interaction between Narration and Reasoning}
\label{tab:cross}
\begin{tabular}{@{}ccccccc@{}}
\toprule
\multirow{2}{*}{Method} & \multicolumn{3}{c}{Narration}            & \multicolumn{3}{c}{Reasoning}            \\ \cmidrule(l){2-4} \cmidrule(l){5-7} 
                               & B4  & C  & M  & B4  & C  & M  \\ \midrule
w/o cross attn            & 32.7 & 234.8 & 30.0 & 10.7 & 96.6 & 15.1 \\ 
w/ swapped attn              & 28.3 & 180.4 & 28.7 & 9.3  & 97.7 & 14.3 \\ \midrule
Ours                  &
  \textbf{34.6} &
  \textbf{247.5} &
  \textbf{30.6} &
  \textbf{11.4} &
  \textbf{102.6} &
  \textbf{15.2} \\ \midrule
Narration only                       & 32.9 & 240.4 & 29.6 & -      & -      & -            \\
Reasoning only                    & -      & -      & -       & 8.1 & 94.4 & 13.3 \\ \bottomrule
\end{tabular}
\end{center}
\end{table}

\subsection{Ablation Study}
We conduct a comprehensive ablation study to analyze various aspects of ADAPT design. 

\textbf{Effect of Action-aware Joint Training}
To investigate the effect of action-awareness in joint training on ADAPT, we train a single captioning model by removing the CSP (control signal prediction) head of ADAPT, referred to as "Single". As shown in Table~\ref{tab:multitask}, ADAPT outperforms single training with an improvement of 15.9 for narration and 7.2 for reasoning on CIDEr metric. This suggests that cues from the other task help regularize the shared video representation and improve the performance of the text generation task.

Additionally, we can see from Fig.~\ref{fig:main}(a) that the caption and control signal data are employed in two streams in ADAPT. An interesting question is: can we simply pass the control signals to the multi-modal transformer to get the final caption prediction?
So we create such an architecture that takes video tokens, control signal tokens (generated by a learnable embedding layer) and masked text tokens as input and generates predictions of the masked tokens, which is referred to as "Single+". Results are shown in the second row of Table~\ref{tab:multitask}. We can see that the proposed ADAPT still achieves the best results, especially for reasoning segment, which demonstrates the superiority of multi-task learning over using both videos and control signals as inputs despite the latter is an intuitive setting.

\begin{table}[]

\begin{center} 
\caption{Impact of Video Frames($T$)}
\label{tab:frame number}
\resizebox{1.0\linewidth}{!}{
\begin{tabular}{@{}cccccccccc@{}}
\toprule
\multirow{2}{*}{Method} & \multicolumn{4}{c}{Narration}                              & \multicolumn{4}{c}{Reasoning} & \multirow{2}{*}{Cost(min)}         \\ \cmidrule(l){2-5} \cmidrule(l){6-9} 
                        & B4  & C           & M & R           & B4  & C  & M & R  \\ \midrule
2                  & 33.4 & 227.7          & 28.7 & 61.0          & 8.7 & 62.9 & 15.1 & 29.8 & 294  \\
4                  & 32.9 & 225.7          & 29.0 & 60.9          & 9.9 & 81.3 & 14.9 & 31.1 & 382 \\
8                  & 32.6 & 236.1          & 29.3 & 61.8          & 8.4 & 83.7 & 13.4 & 30.6 & 447  \\
16                 & 32.5 & 231.0 & 29.5 & 61.9  & 8.7 & 91.5 & 13.8 & 32.0 & 528  \\
32 &
  \textbf{34.6} &
  \textbf{247.5} &
  \textbf{30.6} &
  \textbf{62.8} &
  \textbf{11.4} &
  \textbf{102.6} &
  \textbf{15.2} &
  \textbf{32.0} & 
  797 \\ \bottomrule
\end{tabular}}
\end{center}
\vspace{-0.4cm}
\end{table}

\textbf{Impact of Different Control Signal Types}
In our implementations, we leverage control signals (e.g., course) as supervision for the CSP task. In this analysis, we investigate the impact of different supervision signal types of ADAPT.
The base signals in our experiments are speed and course. We first conduct experiments by removing one of them, results of which are shown in the first two rows of Table~\ref{tab:signal type}. Then in the third row both speed and course are utilized, which is the same as previous experiments. We observe that the removal of each signal leads to the decrease of performance. For example, the CIDEr metric decreases by 29.3 for narration and by 14.0 for reasoning without the speed inputs. This is understandable because being aware of speed and course can help the network learn representations that are informative for narration and reasoning and the lack of either can result in the bias of video representations.

\begin{table*}[]
\begin{center}
\caption{Comparison on Control Signals Prediction Accuracy}
\label{tab:sensor accuracy}
\begin{tabular}{@{}ccccccccccccc@{}}
\toprule
\multirow{2}{*}{Method} & \multicolumn{6}{c}{Course}                               & \multicolumn{6}{c}{Speed}                    \\ \cmidrule(l){2-13} 
                        & RMSE(degree)$\downarrow$ & $A_{0.1}\uparrow$ & $A_{0.5}\uparrow$ & $A_{1.0}\uparrow$         & $A_{5.0}\uparrow$ & $A_{10.0}\uparrow$ & RMSE(m/s)$\downarrow$ & $A_{0.1}\uparrow$ & $A_{0.5}\uparrow$ & $A_{1.0}\uparrow$         & $A_{5.0}\uparrow$ & $A_{10.0}\uparrow$ \\ \midrule
Single              & \textbf{6.3} & 8.3   & 84.7  & \textbf{90.5} & 97.2  & 98.7   & 3.4       & 5.0   & 25.5  & 37.8  & 86.8  & 98.7   \\
Ours &
  6.4 &
  \textbf{62.2} &
  \textbf{85.5} &
  89.9 &
  \textbf{97.2} &
  \textbf{98.8} &
  \textbf{2.5} &
  \textbf{11.1} &
  \textbf{28.1} &
  \textbf{45.3} &
  \textbf{94.3} &
  \textbf{99.5} \\ \bottomrule
\end{tabular}
\end{center} 
\vspace{-0.2cm}
\end{table*}

\textbf{Interaction between Narration and Reasoning}
Compared with the general caption task, the driving caption task generates two sentences: action narration and reasoning. In this section, we explore how these two segments interact with each other by controlling the attention mask or the order of two sentences. 

Specifically, as shown in the right of Fig.~\ref{fig:main}(c), we use a causal self-attention mask for each sentence where a word token can only attend to the existing output tokens, and employ sparse attention \cite{lin2022swinbert} for video tokens. The reasoning segment has full attention to the narration segment, referred to as cross attention, which defines the dependence of reasoning on narration. In this section, we first conduct experiments without cross attention or with swapped cross attention (by swapping the order of narration and reasoning). Results are reported in Table~\ref{tab:cross}. Compared with the default setting (denoted as "Ours"), results without cross attention have lower performance in both sentences, which indicates that conditioning the reasoning segment on the narration segment is beneficial for training. And the performance with swapped cross attention also decreases, especially for the narration part, which further demonstrates this dependence of reasoning on narration, instead of the other way around.

Additionally, we conduct experiments with only one sentence, referred to as "Narration only" and "Reasoning only". Table~\ref{tab:cross} shows that training with both sentences yields improvement on the performance, especially for the reasoning segment, indicating that the interaction between narration and reasoning promotes each component of the full caption task.

\textbf{Impact of Different Sampling Rates}
In previous experiments, we uniformly sample $T=32$ frames from a given video, along with control signal data of the same timestamp. In this study, we investigate the impact of sampling rate by varying the number of sampled frames. Specifically, we uniformly sample $T={2,4,8,16,32}$ frames from a variable-length video, as shown in Table~\ref{tab:frame number}. The performance of ADAPT improves steadily as the sampled number increases since more frames lead to less missing visual content. This suggests that caption results can be enhanced by densely sampled frames and control signals. The training time costs are also provided in Table~\ref{tab:frame number}. We hope this ablation provides robotics practitioner with insights about the accuracy-efficiency trade-off of driving caption.

\begin{figure}
  \centering
  \includegraphics[width=0.47\textwidth]{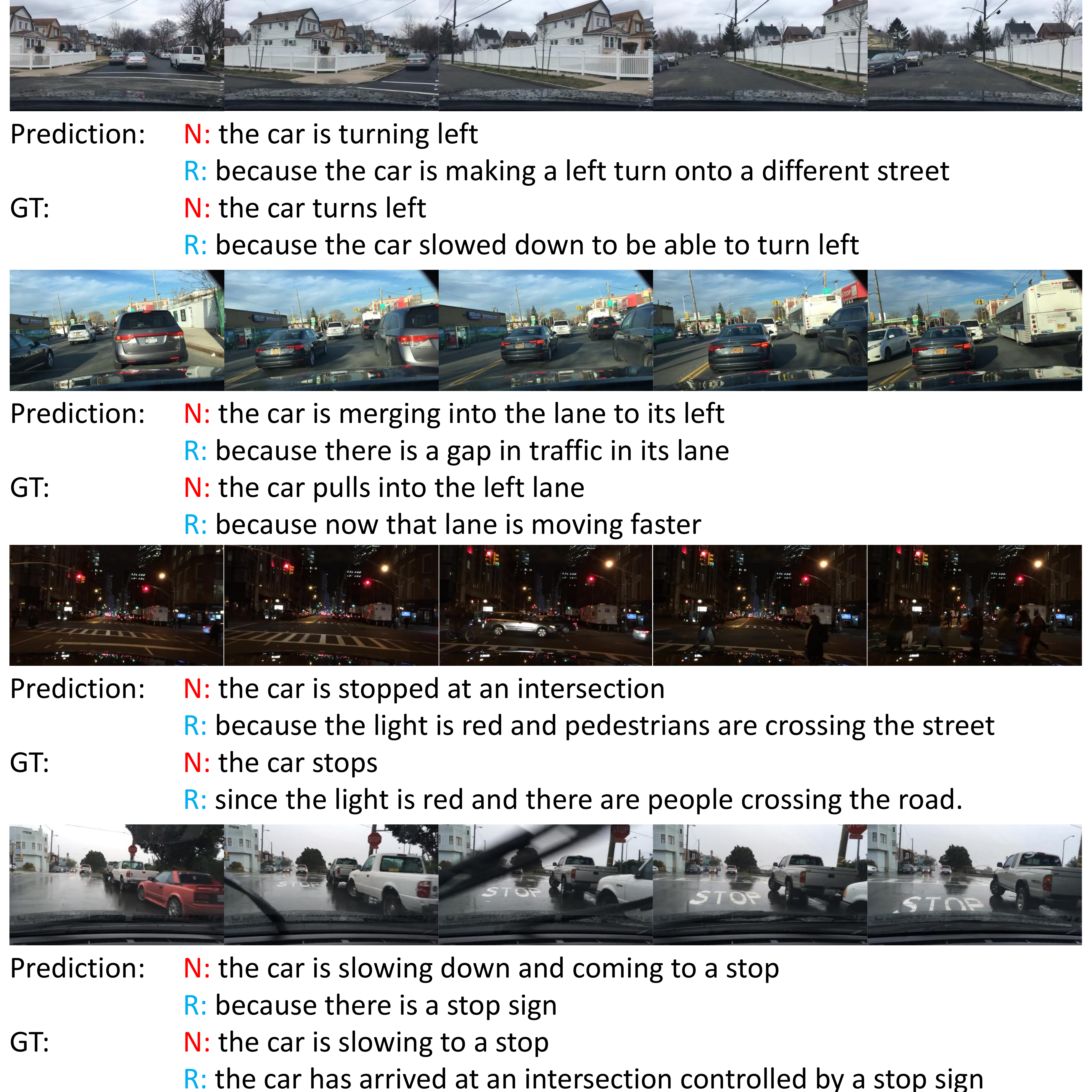}
  \caption{Qualitative analysis: generated narrations correctly describe the current action, with sound reasoning.}
  \label{fig:qualitative}
\end{figure}

\subsection{Analysis on Control Signal Prediction}
Although the main goal of driving caption task is to generate sentences, we also investigate the performance of control signal prediction tasks. We employ root mean squared error (RMSE) and a tolerant accuracy ($A_{\sigma}$) to measure the final performance. Tolerant accuracy means we first use two thresholds to determine the range of the control signal deviation and truncate it. For example, we define the truncation value of predicted course $\hat{c}$ as:
\begin{equation}
    c_{\sigma}=\left\{
        \begin{array}{rcl}
        1,      &      & -\sigma < \hat{c}-c < \sigma\\
        0,      &      & otherwise
        \end{array}
    \right.
\end{equation}
where $c$ is the ground-truth course and $\sigma$ is the tolerant threshold value. Then $A_{\sigma}$ of course represents the accuracy of $c_{\sigma}$ recorded as a percentage, and $A_{\sigma}$ of speed is defined similarly.
Results are provided in Table~\ref{tab:sensor accuracy}. We observe that our joint training framework can further improve the performance of control signal prediction, indicating the benefit of joint training.

\subsection{Deployment in Autonomous Systems}
We further develop a pipeline for the deployment of ADAPT in both the simulator environment (e.g., Carla \cite{dosovitskiy2017carla}) and the real world. The system takes raw vehicular videos as input and generates action narrations and reasoning explanations in real time. Specifically, we first record the frames captured by the camera from the front view. Then the frames in the last several seconds are passed as input to ADAPT to generate the action narration and reasoning of the current step. 
Moreover, we further utilize text-to-speech technology to convert the generated sentences into speech narration, to make it more convenient and more interactive for common passengers (especially helpful for vision-impaired passengers).

\section{Conclusion}
Language-based interpretability is essential for the social acceptance of self-driving vehicles. We present Adapt (Action-aware Driving cAPtion Transformer), a new end-to-end transformer-based framework for generating action narration and reasoning for self-driving vehicles. ADAPT utilizes multi-task joint training to reduce the discrepancy between the driving action captioning task and the control signal prediction task. Experiments on BDD-X dataset over standard captioning metrics as well as human evaluation demonstrate the effectiveness of ADAPT over state-of-the-art methods. We further develop a deployable pipeline for the application of ADAPT in both simulator environment and the real world.

\bibliographystyle{IEEEtran}
\balance
\bibliography{ref}

\end{document}